\documentclass[letterpaper,10 pt, conference]{ieeeconf}  

\IEEEoverridecommandlockouts                              

\usepackage{graphicx} 
\usepackage{subcaption}
\usepackage{caption}

\usepackage{epsfig} 
\usepackage{mathptmx} 
\usepackage{comment}

\makeatletter
\let\NAT@parse\undefined
\makeatother

\usepackage{amsmath} 

\usepackage{amssymb}  
\usepackage{pgfplots}

\usepackage{algorithm}
\usepackage{algpseudocode}
\usepackage{tabularx}
\usepackage{url}
\usepackage{amssymb}
\usepackage{csquotes}
\usepackage{amsmath}

\usepackage{cleveref}
\usepackage{cite}
\usepackage{etoolbox}
\makeatletter
\patchcmd{\@makecaption}
  {\scshape}
  {}
  {}
  {}
\makeatother

\usepackage[letterpaper,bindingoffset=0.2in,%
left=0.75in,right=0.75in,top=0.75in,bottom=0.75in,%
footskip=.25in]{geometry}

\usepackage{accents}

\usepackage{pgfplots}
\usepackage{lipsum}
\usepackage{xcolor}

\title{
\LARGE \bf Qualitative Prediction of Multi-Agent Spatial Interactions 

\author{Sariah Mghames$^1$, Luca Castri$^1$, Marc Hanheide$^1$, Nicola Bellotto$^{1,2}$
\thanks{\noindent\textsuperscript{1}School of Computer Science, University of Lincoln, UK.\newline
\textsuperscript{2}Dept. of Information Engineering, University of Padua, Italy.\newline
This project has received funding from the EU's Horizon 2020 Research and Innovation programme under grant agreement No 101017274}
}
}

\begin{document}
\maketitle
\thispagestyle{empty}
\pagestyle{empty}

\begin{abstract}
Deploying service robots in our daily life, whether in restaurants, warehouses or hospitals, calls for the need to 
reason on the interactions happening in dense and dynamic scenes. In this paper, we present and benchmark three new approaches to model and predict multi-agent interactions in dense scenes, including the use of an intuitive qualitative representation. 
The proposed solutions take into account static and dynamic context to predict individual interactions. They exploit an input- and a temporal-attention mechanism, and are tested on medium and long-term time horizons. 
The first two approaches integrate different relations from the so-called Qualitative Trajectory Calculus~(QTC) within a state-of-the-art deep neural network to create a symbol-driven neural architecture for predicting spatial interactions. 
The third approach implements a purely data-driven network for motion prediction, the output of which is post-processed to predict QTC spatial interactions.
Experimental results on a popular robot dataset of challenging crowded scenarios show that the purely data-driven prediction approach generally outperforms the other two. 
The three approaches were further evaluated on a different but related human scenarios to assess their generalisation capability.
\end{abstract} 


\section{Introduction} \label{sec:intro}
While service robots are increasingly being deployed in our domestic, healthcare, warehouse, and transportation environments, modeling and predicting the {\em interactions} of different agents given their context (i.e. nearby static and dynamic objects, including people) are important requirements for effective human-robot co-existence and intent communication. They can help robots know when, where and how to intervene on the environment, in addition to navigate it safely which is usually accomplished with the classic human motion prediction paradigm. For example, an assistive robot patrolling someone's home or a crowded hospital needs to reason continuously on the relative state of nearby people for approaching and communicating with them, ideally predicting future human (spatial) interactions to optimize its own decision-making.

\begin{figure}[t]\centering
\includegraphics[trim={330 160 310 100},clip, scale=0.55]{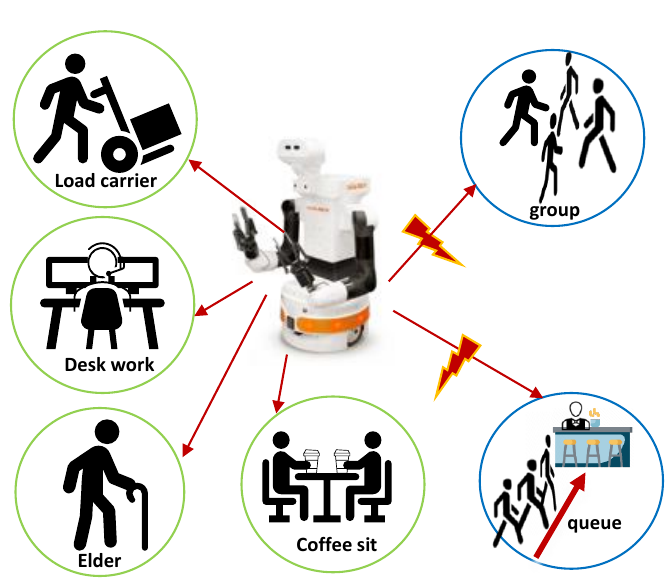}
\caption{Modeling and predicting multi-agent interactions can improve the decision-making process of social and service robots for helping in heavy and/or health-related tasks, anticipating elders assistance needs, cafe/restaurant table serving, office duties, etc., while avoiding conversational groups or ordering queues (unless called for it).}\label{fig:intro}
\end{figure}

Differently from typical human motion prediction, which is mostly concerned with navigation safety, dealing with multi-agent {\em interactions}
presents two advantages: from a ``social'' navigation point of view, 
interaction prediction facilitates human-robot motion coordination and intent communication; from an ``explainable'' decision-making point of view, interaction prediction makes an individual's motion behaviour more meaningful in many social contexts.
For example, by detecting a group meeting in an office (as in~Fig.\ref{fig:intro}), and predicting it will last for a while, the robot does not disturb the people involved. On the other hand, if the robot predicts that an elderly patient is trying to approach it to ask for help, it can use this information to update its initial plan and prioritize the responsiveness to the human's intent. 

An intuitive approach for multi-agents spatial interaction representations is the qualitative one. In the 2D and 3D spatial domains (e.g. navigation and manipulation, respectively), a qualitative interaction between pairs of agents or body points can be captured by some symbolic representation. One way to model qualitative spatial interactions is by using a qualitative trajectory calculus~(QTC)~\cite{delafontaine2012qualitative}. QTC-based models of moving agent pairs can be described by different combinations of QTC symbols that represent spatial relations, e.g. relative distance (moving towards/away), velocity (faster/slower), or orientation (to the left/right) with respect to the central axis joining both agents.
QTC-based interaction modeling was presented in~\cite{bellotto2012robot,hanheide2012analysis,dondrup2014probabilistic} for modeling 2D human-robot spatial interactions, with further application into human-aware robot navigation~\cite{bellotto2013qualitative,dondrup2016qualitative}. Differently from the focus of this paper, the authors in~\cite{dondrup2016qualitative} used a Bayesian temporal model to study the interaction of a single pair of agents (human-robot), without accounting for the dynamic and/or static context, which limits the prediction performance.
An alternative way for representing spatial interactions in a multi-agent scenario is by quantitatively merging all agents in the context to drive a robot navigation stack~\cite{chen2019crowd}-\cite{pokle2019deep}. 
These works though cannot infer the implicit spatials intent of the agents. 


To the best of our knowledge, there is a gap in the literature regarding the prediction of qualitative (i.e symbolic) and/or quantitative (i.e metrical) interactions between multi-agent entities (e.g. human-human, human-robot, human-object, robot-object) given their nearby dynamic and/or static context, which was only partly addressed in~\cite{dondrup2016qualitative} for a single human-robot pair. Further investigation in more complex scenarios is therefore necessary to enhance future robot reasoning, mutual intent communication, and reactive or predictive planning. 

The contribution of this paper is therefore two-fold: (i)~addressing the prediction of Multi-Agent Spatial Interactions~(MASI) with dynamic and static context-awareness by implementing three new approaches for medium and long-term interactions predictions, including a QTC-based neural network representation;
(ii) experimentally evaluating the proposed frameworks on different scenarios with multiple humans/objects to assess the prediction performance, even under domain-shift conditions.

The remainder of the paper is as follows: Sec.~\ref{sec:lit} presents an overview of the related work; Sec.~\ref{sec:appr} explains the approach adopted to model and predict spatial interactions in dense scenes; Sec.~\ref{sec:exp} illustrates and discusses the results from experiments conducted on a public dataset for social robot navigation; finally, Sec.~\ref{sec:conc} concludes by summarising the main outcomes and suggesting future research work.

\section{Related Work} \label{sec:lit}
\textbf{Human-human interactions modeling:}
Two methods have been presented in the literature for interactions modeling with nearby dynamic agents: (i) one-to-one modeling, and (ii) crowd modeling. A one-to-one interaction modeling between human-robot pair was presented in~\cite{dondrup2014probabilistic} in the form of qualitative representation by encoding a sequence of QTC states in a Markov Chain representation. Human-human interactions modeling was also addressed in~\cite{chen2019crowd} for social navigation, where interactions with neighbors are embedded in a multi-layer perceptron by using local maps centered at each person. 
On the other hand, crowd modeling was discussed in~\cite{Ijaz2015}, where the major existing hybrid techniques were surveyed. Hybrid crowd techniques are brought forward to overcome some limitations of classical methods (e.g high computation cost). 
For crowd analysis, F-formations modeling and detection has been addressed recently in~\cite{hedayati2020reform}, where 
the authors deconstructed a social scene into pairwise data points, then they used feature-based 
classification to distinguish F-formations. 
In this work, we do not limit our approach to F-formations only. Hence, we build on previous works from HRSI modeling for single pair of agents~\cite{dondrup2016qualitative}, taking inspiration from the hybrid approaches of crowd modeling, in order to predict multi-agent interactions in dense scenes.
\newline
\textbf{Context-aware human motion prediction:} 
While the problem of context-aware human motion prediction has been extensively addressed in the literature, to the best of our knowledge, the problem of context-aware multi-agent interactions prediction in dense environments (as the social ones) has been mostly neglected. The state of the art works vary based on no context-awareness, static-only context,
dynamic-only context~\cite{Alahi2016,gupta2018social,huang2019stgat}, static and dynamic context~\cite{tao2020dynamic,liang2019peeking}. 
\newline
Architectures such as Social-LSTM~\cite{Alahi2016} and SGAN~\cite{gupta2018social} capture spatial interactions only. Also, the authors adopt an LSTM encoding for each agent that cannot account for static objects embedding in the neighborhood. The Stgat architecture in~\cite{huang2019stgat} 
accounts for dynamic agents only and the use of dual LSTM modules in the encoder limits the ease of direct integration of static objects representations. As per~\cite{tao2020dynamic}, the DSCMP architecture outperforms S-LSTM, SGAN and Stgat in terms of parameters and time consumption. In DSCMP, both static and dynamic contexts are incorporated together with spatial and temporal dependencies between agents. In that work, the static context is embedded in a latent space through a convolutional semantic map of the whole scene. 
The work in~\cite{liang2019peeking}, instead, addresses the problem of action prediction together with motion prediction, by using person-whole scene interaction embedding (leveraging the semantic scene convolutional map) together with explicitly encoding the interaction between person and surrounding objects (person, objects) into a geometric relationship.

In this paper, we take inspiration from~\cite{tao2020dynamic} and~\cite{liang2019peeking}
to develop a dynamic and static context-aware predictor of spatial interactions, but we limit our current study to one data type entry to the network architecture used for experimentation. We choose raw coordinates as the sole upstream data type, commonly used to represent dynamic agents motion, leaving the exploitation of semantic map representation of the scene (fully or partially) for our future work. Hence, we embed only the raw coordinates of key features (static objects of use) that represent the social scene, and that's because in social scenes and according to~\cite{manzo2014machines}, humans interact not only with one another, but also with machines that are meaningful.  
\section{MASI Prediction Framework} \label{sec:appr}
\subsection{Problem Statement}
While metrical motion prediction of nearby agents allows robots to replan locally their target destination for safe navigation, it doesn't provide the robot with enough intelligence to reason on the implicit intent one may convey in his motion (e.g. a person may speed up at a room entrance, as a patient's room, to convey to the robot an urgent need to enter first), the problem of which can be dealt by embedding the robot with a reasoning (modeling and predicting) paradigm on multi-agent spatial interactions, allowing it to take reactive or predictive optimal decisions by intervening or not on its surrounding.

\subsection{Qualitative Spatial Interactions} \label{sec:qtcform}
\begin{figure}[h]\centering
\includegraphics[trim={250 225 420 170},clip, scale=0.6]{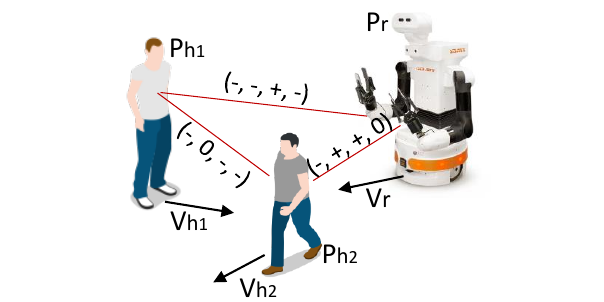}
\caption{A case of $QTC_{C_1}$ representation of interactions between three body points $P_{h1}$, $P_{h2}$, and $P_r$. For example, the QTC interaction represented by $(-, -, +, -)$ implies that agent $h_1$ and robot `r' are moving towards each other, `r' moves to $h_1$ right side, while $h_1$ moves to `r' left side.}\label{fig:qtcc}
\end{figure}

A qualitative spatial interaction is represented by a vector of $m$ QTC symbols ($q_i$, $i \in \mathbb{Z}$) in the domain $D=\{-, 0, +\}$~\cite{delafontaine2012qualitative}. Among the different types of QTC representations, we exploit the double-cross $QTC_C$ which employs away/towards, left/right, relative speed, and relative angle dichotomies, as illustrated in Fig.~\ref{fig:qtcc}. Two types of $QTC_{C}$ were proposed in the literature, 
the $QTC_{C_1}$ with four symbols $\{q_i, \, i = 1..4\}$, and the $QTC_{C_2}$ with six symbols $\{q_i, \, i = 1..6\}$. The symbols $q_1$ and $q_2$ represent the towards/away (relative) motion between a pair of agents; $q_3$ and $q_4$ represent the left/right relation; $q_5$ indicates the relative speed, faster or slower; finally, $q_6$ depends on the (absolute) angle with respect to the reference line joining a pair of agents. The qualitative interaction between the time series of two moving points, $P_r$ and $P_h$, is expressed by the following $q_i$ symbols: 

\begin{footnotesize}
\centering
\begin{align*}
(q_1)  ~ &- : d(P_r|t^-, P_h|t) > d(P_r|t, P_h|t) \\
 &+: d(P_r|t^-, P_h|t) < d(P_r|t, P_h|t) ; \quad 0: \text{all other cases}\\
(q_3) ~ &-: \|\vec{P_r^{t^+}P_r^t} \wedge \vec{P_h^tP_r^t}\| < 0 \\
 &+: \|\vec{P_r^{t^+}P_r^t} \wedge \vec{P_h^tP_r^t}\| > 0 ; \quad 0: \text{all other cases} \\
(q_5) ~ &-: \|\vec{V_r^t}\| < \|\vec{V_h^t}\| \\
&+: \|\vec{V_r^t}\| > \|\vec{V_h^t}\| ; \quad 0: \text{all other cases} \\
(q_6) ~ &-: \theta(\vec{V_r^t}, \vec{P_rP_h}^t) < \theta(\vec{V_h^t}, \vec{P_hP_r}^t) \\
 &+: \theta(\vec{V_r^t}, \vec{P_rP_h}^t) > \theta(\vec{V_h^t}, \vec{P_hP_r}^t) ; \quad 0: \text{all other cases}
\end{align*}
\end{footnotesize}

\noindent ($q_2$) and ($q_4$) are similar to ($q_1$) and ($q_3$), respectively, but swapping $P_r$ and $P_h$. $d(.)$ is the euclidean distance between two positions, $\theta(.)$ is the absolute angle between two vectors, and $t^-$ denotes a single previous time step.

In this paper, we propose a framework ($F$) for spatial interaction prediction and compare three possible implementations: two symbol-driven neural approaches, denoted by $F^{QTC-4}$ and $F^{QTC-6}$, where both input and output of the neural network are QTC symbols; a third approach, denoted by $F^{ts}$, where the inputs are raw trajectories and the outputs are QTC symbols. In particular, $F^{QTC-4}$ and $F^{QTC-6}$ exploit $QTC_{C_1}$ and $QTC_{C_2}$, respectively, to directly predict QTC vectors with a time horizon $T_f$, while $F^{ts}$ extracts QTC vectors from the coordinates generated by a purely data-driven motion prediction architecture over $T_f$. The main difference between the two symbol-driven frameworks is that $F^{QTC-4}$ assigns a greater importance
to the prediction of left/right and towards/away dichotomies, neglecting the relative velocity and angle embedded in~$F^{QTC-6}$.

\begin{figure*}[t]\centering
\includegraphics[trim={100 210 70 173},clip, scale=0.55]{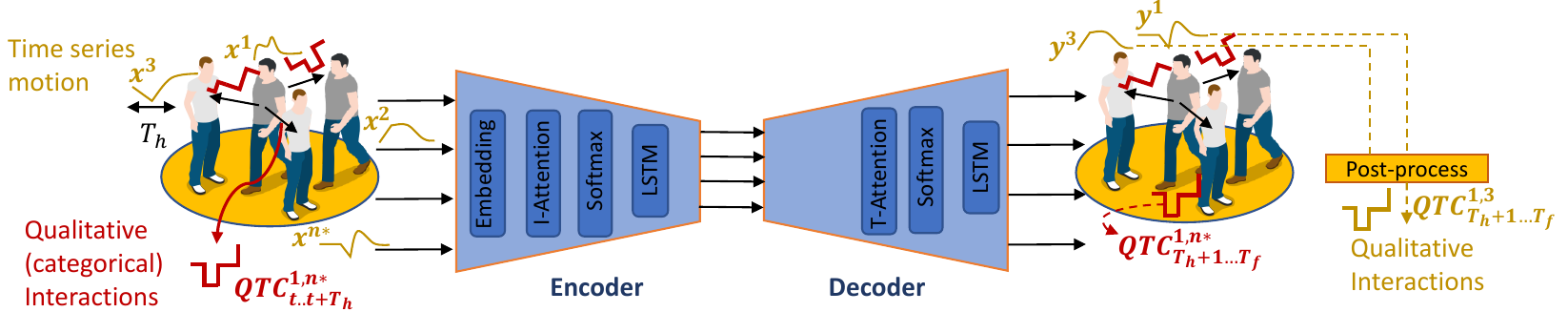}
\caption{An input-temporal attention mechanism for predicting spatial interactions of multi-dimensional input categorical (red) and metrical (yellow) time series extracted from dense scenes: application to JackRabbot dataset. The diagram is extended from~\cite{qin2017dual}. $\mathbf{x}$ is the input driving vector, $\mathbf{y}$ is the label vector.}\label{fig:network}
\end{figure*}

\subsection{Network Architecture} \label{sec:net}
In order to narrow down the study, we limit the network upstream input to raw coordinates of body points (i.e. dynamic agents, static key objects in the environment), which will then be converted to QTC input for the evaluation of $F^{QTC-4}$ and $F^{QTC-6}$ frameworks. Among several network architectures developed in the literature for human motion prediction, some give no consideration to the static context~\cite{gupta2018social}, others embed the static context as semantic map input to the network~\cite{tao2020dynamic,liang2019peeking}. Though these architectures can serve as a tool for our current study, we take advantage in this paper from the network architecture in~\cite{qin2017dual} as starting point to implement our interaction prediction framework (F) for the prediction of qualitative spatial interactions. The architecture as in~\cite{qin2017dual} takes as input only time series of raw coordinates which get processed through an embedding, attention, and LSTM layers respectively, as can be seen from the encoder and decoder in Fig.~\ref{fig:network}. This architecture alleviates the need for a separate network for static context embedding, as a CNN features extractor from the semantic scene image. The architecture in~\cite{qin2017dual} allows also the incorporation of both spatial and temporal dependencies of interactions. It is worth to stress on the fact that other architectures from the state of the art in context-aware human motion prediction can serve the purpose of this benchmark study, and this will be targeted in our future works for performance generalisation. 
In order to implement $F^{QTC-4}$ and $F^{QTC-6}$, we modified the original architecture to deal with time-series of categorical data, representing symbolic knowledge of the spatial interactions between pairs of agents. We also extended the prediction horizon to medium (i.e. 48 time steps, or $3.2s$) and longer (i.e. 72 time steps, or $4.8s$) time horizons. The parameters for medium and long time horizon prediction were chosen based on relevant literature of human motion prediction~\cite{gupta2018social,tao2020dynamic}.
The input attention encoder of the network in Fig.~\ref{fig:network} consists of an input attention layer (I-Attention) which weighs $n^*$ spatial interactions in a radial cluster. The encoder is then followed by a decoder with a temporal attention layer (T-Attention), capturing the temporal dependencies in multi-agent interactions. The network encodes $n^*$ input series (denoted by $\mathbf{x}$), each of length $T_h$, and decodes $n^*$ output labels (denoted by $\mathbf{y}$), each of length $T_f$, where $T_f$ is the predictive time horizon and $T_h$ is the time history used for the temporal attention purpose. 
For our categorical data, we minimize a sparse (categorical) cross-entropy loss function between the true and predicted QTC vector indices, extracted from a dictionary of 444 possible $QTC_{C_2}$ vectors for $F^{QTC-6}$, and 82 possible $QTC_{C_1}$ vectors for $F^{QTC-4}$. Both the dictionaries include an additional index of ``impossible'' QTC vector, where all the QTC relations $q_i$ assume a value $\notin D$, chosen to be $10$. 
The impossible QTC vector accounts for the case of an agent leaving the cluster at time $t$, or for complementary ``fake'' interactions added to each cluster to make it of fixed $n^*$ size. 
The reader can refer to~\cite{qin2017dual} for a detailed explanation of the network components, which are schematically illustrated in Fig.~\ref{fig:network}.

\begin{figure}[t]\centering
\includegraphics[scale=0.045]{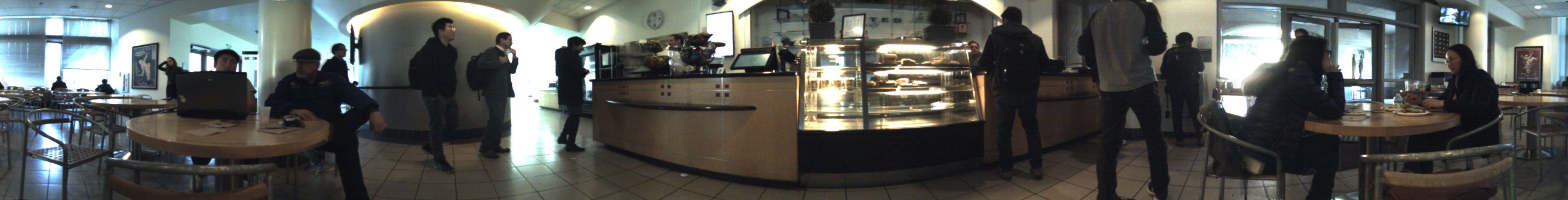}\\
\vspace{2pt}
\includegraphics[scale=0.045]{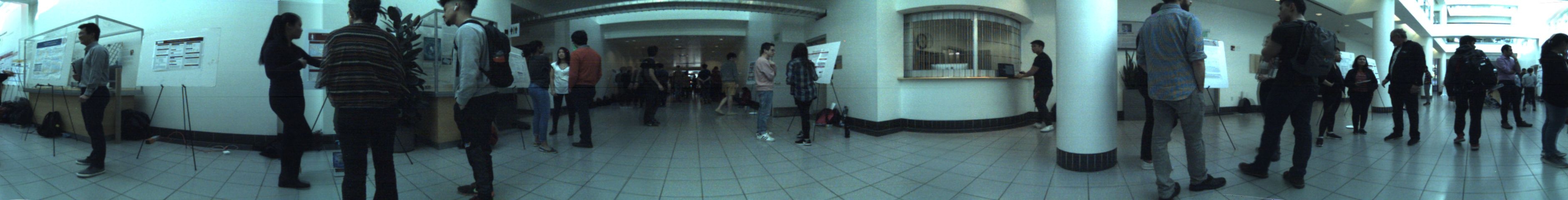}
\caption{JackRabbot dataset scenes: (top) \textit{bytes-cafe-2019-02-07\_0}, (bottom) \textit{packard-poster-session-2019-03-20\_2}.}\label{fig:dataset}
\end{figure}

\subsection{Data Processing} \label{sec:data_processing}
Social scenarios have often an unpredictable number of people entering  and leaving the environment, possibly leading to a combinatorial explosion in the input size of the predictive model and in its number of training parameters. 
In order to approach the problem of reasoning in socially crowded environments, we implement a crowd clustering approach for local interactions prediction. The advantage of this approach is that all the clusters have a fixed micro-size (i.e maximum number of agents entering the cluster at any given time) and it accounts for the agents entering and leaving the cluster.
We applied the radial clustering approach on the JackRabbot (JRDB~\footnote{https://jrdb.erc.monash.edu/})
open-source dataset, which provides multi-sensor data of human behaviours from a mobile robot in populated indoor and outdoor environments (Fig.~\ref{fig:dataset}). We make use of the open-source annotated 3D point clouds, provided as metric coordinates of humans (dynamic context) bounding boxes centroid, and extracted from the upper velodyne sensor, as raw data and ground truth to our network architecture.
The raw data are further processed to extract QTC representations of a spatial interaction between each pair of agents, whose dictionary index is then used as ground truth output for $F^{QTC-4}$ and $F^{QTC-6}$ approaches. In parallel, the raw metric data are directly used as ground truth labels for the $F^{ts}$ approach. The environments considered in JRDB are fairly crowded. Among them, we selected a cafe shop (\textit{bytes-cafe-2019-02-07\_0}) for comparing the proposed prediction approaches, and two poster session scenarios (\textit{packard-poster-session-2019-03-20\_2}, denoted PS-2, and \textit{packard-poster-session-2019-03-20\_1}, denoted PS-1) for testing the framework on a domain-shift situation.
In the cafe scenario, the static context includes objects such as bar order and check-out points, exit door, drinking water station, as illustrated in Fig.~\ref{fig:dataset} (top). These objects were manually selected based on our careful investigation to identify the most common ones used by people in the scenario, although in the future we plan to learn them automatically in order to adapt to different environments. The spatial coordinates of the selected objects, extracted from the scene point cloud, are incorporated in the network architecture as any other dynamic agent.
\newline
 For each agent $i$ in a given scene, we generate a cluster with a fixed interaction radius $R_1=1.2m$. The latter is selected based on the proxemics' literature~\cite{hall1969hall}, where the social distance for interactions among acquaintances is indeed between $1.2m$ (short phase) and $3.7$m (long phase). Each cluster includes $n$ input series, with $n$ being the maximum number of agents entering the cluster of agent $i$ in a time interval $T$. Each input series is defined as a series of spatial interactions between agents $r$ and $h$, where $h$ is every other dynamic agent/static object in the cluster of $r$. The maximum number of input series among all clusters, $n^*$, is fixed for practical (training) purposes. 
 Each cluster is then post-processed to include $(n^* - n)$ input series with complementary ``fake'' values.
 
 Spatial interactions are formulated in terms of categorical (i.e. QTC) data, hence two dictionaries of all possible qualitative interactions in 2D are generated based on $QTC_{C_2}$ and $QTC_{C_1}$ for the approaches $F^{QTC-6}$ and $F^{QTC-4}$, respectively. The input to our network are now $n^*$ series of indices over the time history~$T_h$. For both the cafe and the poster sessions scenarios, we evaluated the prediction performance for a medium ($T_f = 3.2s$) and a longer term ($T_f = 4.8s$) horizons.

\section{Experiments} \label{sec:exp}
The three proposed framework configurations implement the same architecture as in Fig.~\ref{fig:network} but they were trained with different losses, since the input data is different.
$F^{QTC-4}$ and $F^{QTC-6}$ were trained by minimising a categorical cross-entropy loss function over 120 epochs using Adam optimiser, and with $T_h = 10$ time steps (i.e $0.67s$, much less than other works as in~\cite{gupta2018social}, \cite{huang2019stgat}), a batch size $B = 10$, as hyper-parameters, while $F^{ts}$ was trained using the root mean square error loss function (RMSE) over 80 epochs using Adam optimiser, and with $T_h = 5$ time steps and $B = 5$. Other common hyper-parameters between the 3 network configurations are, hidden states $h = 256$ for both the encoder and decoder and a learning rate $l_r = 0.001$. The hyper-parameters were tuned to reach a good validation loss. The input consists of $63,566$ samples for the cafe scene with $F^{QTC-4}$ and $F^{QTC-6}$, and $46,548$ with $F^{ts}$; $109,040$ samples for PS-2 with $F^{QTC-4}$ and $F^{QTC-6}$, and $110,889$ with $F^{ts}$, whereas PS-1 has $69,126$ samples in the three frameworks. The size of the input dataset is the same for both medium and longer term $T_f$, and it is divided into $80\%$ training, $10\%$ validation, and $10\%$ testing sets. All the three frameworks were trained on a computing system consisting of Intel® Core™ i7-6850K processor $@3.6GHz$ and NVIDIA GeForce GTX 1080 Ti 11GB GPU.

Since the three proposed approaches for spatial interaction prediction are trained with different loss functions, in order to compare their performance we use the so-called ``conceptual QTC distance''~\cite{delafontaine2012qualitative} defined as a measure for the closeness of QTC relations. Specifically, a conceptual distance between ${0}$ and another symbol, $\{+\}$ or $\{-\}$, is assumed to be \enquote{$+1$}, while the conceptual distance between $\{+\}$ and $\{-\}$ is \enquote{$+2$}. The overall conceptual distance between two QTC vectors is calculated by summing the conceptual distance over all their relation symbols. For example, suppose $QTC^t$ and $QTC^p$ are two QTC vectors, where $t$ and $p$ refer to the true and predicted QTC vectors, respectively. Then, the conceptual QTC distance is calculated as: 
\begin{equation}\label{eq:ddqtc}
\mathbf{d}_{QTC} = \mathbf{d}_{QTC^t}^{QTC^p} = \sum_{q_i} \mid  q_i^{QTC^t} - q_i^{QTC^p} \mid, 
\end{equation}
where $q_i$ is one of the symbols defined in Sec.~\ref{sec:qtcform}.

\subsection{Testing Evaluation}

In Table~\ref{tab:sum}, we report the results on the $10\%$ test set and cluster radius of $R_1$ = $1.2$m of the cafe scene in terms of normalised mean ($\mu$) and standard deviation ($\sigma$) of $d_{QTC}$. The normalisation is done over the labels, $T_f$, and $B$. We note that the range of $d_{QTC}$ is approximately $\mathcal{R}=\{0-40\}$ for $F^{QTC-4}$, and $\{0-60\}$ for $F^{QTC-6}$. The maximum value of $\mathcal{R}$ accounts for the inability of QTC to represent missing agents in the radial cluster.
On the test set, $F^{QTC-6}$ significantly outperforms $F^{QTC-4}$ over medium and longer time horizons, however $F^{ts}$ have the best performance among the three configurations, over both time horizons. Also, $F^{ts}$ (motion prediction) with $QTC_{C_1}$ post-processing (denoted $F^{ts,1}$) for interaction prediction or analysis performs better on the medium term while $F^{ts}$ with $QTC_{C_2}$ ($F^{ts,2}$) performs best on the longer term. Overall, $F^{ts,1}$ and $F^{ts,2}$ outperform $F^{QTC-6}$, with $F^{ts,1}$ having $73.05\%$ and $81.27\%$ reduction on $\mu$($d_{QTC}$), and $93.8\%$ and $96.68\%$ reduction on $\sigma$($d_{QTC}$), over the medium and long term predictions, respectively. From these observations we can conclude that predictive networks perform better on non-symbolic data compared to their symbolic counterpart when applied to crowded human environments. We report $F^{ts,1}$ training and validation time of $6.6hrs$ and $8.3hrs$, while the evaluation time is $5.8ms$ and $9.3ms$
over $3.2s$ and $4.8s$ prediction horizons, respectively.
In order to evaluate the effect of cluster radius selection, Table~\ref{tab:sum} also shows the results when cluster radius is $R_2$ = $3.7$m. In this case, with a larger cluster, hence with more context accounted for, $F^{ts,1}$ outperforms all other configurations, on both the medium and longer horizons, it also outperforms $F^{ts,2}$ performance over $T_f = 4.8s$ and when $R_1$ is accounted for. We can infer that with larger cluster radius, more context is accounted for to help in long term prediction, and hence, less interaction symbols are required to accurately represent the true interactions between multi-agents.

\begin{table}
\begin{center}
\textbf{Cafe}  \\
\vspace{2pt}
\begin{tabular}{l|l|l||l|l}
\hline
\hline
& $\mu^{10\%-R_1}$ & $\sigma^{10\%-R_1}$ & $\mu^{10\%-R_2}$ &  $\sigma^{10\%-R_2}$\\
\hline
$\textbf{F}^{QTC-6}$ (3.2s) & 1.772 & 3.568 & 3.064 & 3.851 \\ 
\hline
$\textbf{F}^{QTC-4}$ (3.2s) &  7.545 & 4.067 & 3 & 3.857\\
\hline
$\textbf{F}^{ts,1}$ (3.2s) &  \textbf{0.464} & \textbf{0.22} & \textbf{0.32} & \textbf{0.16}\\
\hline
$\textbf{F}^{ts,2}$ (3.2s) &  0.68 & 0.166 & 0.638 & 0.11\\
\hline
$\textbf{F}^{QTC-6}$ (4.8s) & 3.44 & 4.4 & 3.46 & 4\\ 
\hline
$\textbf{F}^{QTC-4}$ (4.8s) & 7.61 & 4.057 & 3.8 & 4.18\\
\hline
$\textbf{F}^{ts,1}$ (4.8s) & 3 & 1.254 & \textbf{0.25} & \textbf{0.18} \\
\hline
$\textbf{F}^{ts,2}$ (4.8s) & \textbf{0.644} & \textbf{0.146} & 0.55 & 0.13 \\
\hline
\end{tabular}
\caption{Performance comparison between the QTC prediction approaches $F^{QTC-4}$ and $F^{QTC-6}$, and the motion prediction-based QTC analysis framework $F^{ts}$ evaluated on $QTC_{C_1}$ ($F^{ts,1}$) and $QTC_{C_2}$ ($F^{ts,2}$), in the cafe scene of JRDB and over $T_f$ = $3.2s$ and $4.8s$ prediction horizons. All measures are unitless. $\mu$ and $\sigma$ are the normalised mean and standard deviation of the conceptual distance ($d_{QTC}$) measure over the test set. $R_1$ and $R_2$ correspond to cluster radius $1.2$m and $3.7$m, respectively. The best performance is highlighted in bold.} \label{tab:sum}
\end{center}
\end{table}

\subsection{Domain-Shift (DS) Evaluation}
In order to further assess the generalisation capabilities of the three approaches, we re-trained and compared the results on different but related scenarios. Unfortunately, another cafe scene in JRDB (\textit{forbes-cafe-2019-01-22\_0})
lacks the necessary information to transform local coordinates from a mobile robot into a fixed reference frame for further data processing. Therefore, without loss of generality, we chose another crowded environment (poster session PS-2, as in Fig.~\ref{fig:dataset}-bottom) to re-train our network configurations with $R_1$=$1.2$m, and tested the latter on a different but related scenario (poster session PS-1). 
The performance on the testing set (i.e. $10\%$ of PS-2) is reported in Table~\ref{tab:poster1} (first column). We notice that $F^{ts,1}$ outperforms $F^{QTC-4}$ and $F^{QTC-6}$ on both medium and long term predictions with $72.47\%$ and $85.8\%$ reduction on $\mu$($d_{QTC}$), and $93.9\%$ and $94.48\%$ reduction on $\sigma$($d_{QTC}$), for the $3.2$ and $4.8$s horizons, respectively. We note that, even within the same network configuration $F^{ts,1}$ outperformed $F^{ts,2}$. When looking at the transfer domain PS-1 in Table~\ref{tab:poster1} (second column), on the 100\% dataset, all the configurations succeeded in generalising to PS-1 on the medium and longer terms except $F^{ts,1}$ and $F^{ts,2}$ who generalised well only on the medium term. Nevertheless, $F^{ts,1}$ keeps holding the best performance overall when looking only at PS-1.

In summary, we can infer that, $F^{ts,1}$ is the best framework for developing qualitative predictive solutions to embed a social autonomous system with additional intelligent capabilities as inferring on implicit intent communication and/or predict a need from the surrounding agents. A typical real-world scenario can be a robot patrolling an elderly home care center and instantly inferring on an elder approaching it for requesting assistance in taking a treatment (e.g. bringing water, pills). $F^{ts,1}$
shows lowest mean and standard deviation loss, over short and longer horizons, and among different cluster radius. It also transfers to other domains with a $12.2\%$ decrease and $17.8\%$ increase in mean loss, over $3.2s$ and $4.8s$, respectively.

\begin{table}
\begin{center}
\begin{tabular}{p{1.5cm} p{1.2cm} p{1.2cm} || p{1.2cm} p{1.15cm}  }
   & \textbf{PS-2} & & \textbf{PS-1} \\ 
\hline 
\hline
 & $\mu^{10\%}$ & $\sigma^{10\%}$ &  $\mu^{100\%}$  & $\sigma^{100\%}$ \\
\hline
$F^{QTC-6}$(3.2s) & 1.78 &  3.558 & 0.77 & 0.26\\
\hline
$F^{QTC-4}$(3.2s) & 7.34 &  4.08 & 1.3 & 1.86\\
\hline
$F^{ts,1}$(3.2s) & \textbf{0.49} &  \textbf{0.217} & \textbf{0.43} & \textbf{0.22}\\
\hline
$F^{ts,2}$(3.2s) & 0.715 &  0.17 & 0.7 & 0.17\\
\hline
$F^{QTC-6}$(4.8s) & 2.098 &  3.81 & 0.84 & 0.26  \\
\hline
$F^{QTC-4}$(4.8s) & 8.018 &  3.7 & 1.27 & 1.92  \\
\hline
$F^{ts,1}$(4.8s) & \textbf{0.297} &  \textbf{0.21} & \textbf{0.35} & \textbf{0.22} \\
\hline
$F^{ts,2}$(4.8s) & 0.547 &  0.137 & 0.6 & 0.15 \\
\hline
\end{tabular}
\end{center}
\caption{Performance comparison between the QTC prediction approaches $F^{QTC-4}$ and $F^{QTC-6}$, and the motion prediction-based QTC analysis framework $F^{ts}$ evaluated on $QTC_{C_1}$ ($F^{ts,1}$) and $QTC_{C_2}$ ($F^{ts,2}$), in the poster halls PS-2 and PS-1 of JRDB and over $T_f$ = $3.2s$ and $4.8s$ prediction horizons. All measures are unitless. $\mu$ and $\sigma$ are the normalised mean and standard deviation of the conceptual distance ($d_{QTC}$) measure over $10\%$ test set (PS-2) and $100\%$ test set (PS-1). The best performance is highlighted in bold.} \label{tab:poster1}
\end{table}

\section{Conclusion} \label{sec:conc}
In this work, we presented and compared three approaches for multi-agent prediction of qualitative interactions in dense social scenes, combining a symbolic motion representation with an input/temporal-attention network architecture. We implemented a radial clustering approach to address mainly the notion of social proximity, and formulated spatial interactions in terms of a qualitative trajectory calculus~(QTC). We compared two symbol-driven neural networks for QTC prediction, $F^{QTC-4}$ and $F^{QTC-6}$, with a third purely data-driven approach, $F^{ts}$, based on plain coordinates,
and evaluated them over two fixed-time horizons. We showed that the latter solution outperforms the previous two, specifically when post-processed for a small number of QTC symbols ($F^{ts,1}$), and that it performs best in the domain-shift scenario.
Our future work will be devoted to 
the exploitation of this prediction framework for effective human-robot spatial interactions in social navigation applications, including real-world environments such as warehouses and university premises.
In addition, we will further improve our models to select and integrate learnable key features of the environment, whether static or dynamic, which could have some causal influence on the aforementioned interaction processes.  

\section*{Acknowledgement}
The authors would like to thank Francesco Castelli for his support in designing the problem approach.

\bibliographystyle{IEEEtran}
\bibliography{IEEEabrv,references}

\end{document}